\documentclass[11pt,letterpaper]{article}
\usepackage{amssymb,amsmath}
\usepackage{graphicx}
\usepackage{hyperref}
\usepackage{natbib}
\usepackage{times}
\usepackage{xcolor}

\title{Coupling Speech Encoders with Downstream Text Models}
\author{Ciprian Chelba and Johan Schalkwyk\\
  Google, Inc.\\
  1600 Amphitheatre Parkway\\
  Mountain View, CA 94043, USA\\
  {\tt \{ciprianchelba,johans\}@google.com}}

\date{April 13, 2023}

\begin{document}
\maketitle

\begin{abstract}
We present a modular approach to building cascade speech translation (AST) models that guarantees that the resulting model performs no worse than the 1-best cascade baseline while preserving state-of-the-art speech recognition (ASR) and text translation (MT) performance for a given task. 

Our novel contribution is the use of an ``exporter'' layer that is trained under L2-loss to ensure a strong match between ASR embeddings and the MT token embeddings for the 1-best sequence. The ``exporter'' output embeddings are fed directly to the MT model {\it in lieu} of 1-best token embeddings, thus guaranteeing that the resulting model performs no worse than the 1-best cascade baseline, while allowing back-propagation gradient to flow from the MT model into the ASR components. 

The matched-embeddings cascade architecture provide a significant improvement over its 1-best counterpart in scenarios where incremental training of the MT model is not an option and yet we seek to improve quality by leveraging (speech, transcription, translated transcription) data provided with the AST task. The gain disappears when the MT model is incrementally trained on the parallel text data available with the AST task.

The approach holds promise for other scenarios that seek to couple ASR encoders and immutable text models, such at large language models (LLM).
\end{abstract}

\section{Introduction} \label{sec:intro}

Automatic speech translation (AST) modeling is usually plagued by lack of parallel training data, which limits the success of end-to-end models.  Owing to their modular architecture, cascade models for AST have the advantage of leveraging the large amounts of data available to build automatic speech recognition (ASR) and machine translation (MT) models, respectively. The straightforward way of building cascade AST models is to send the 1-best ASR transcription to the text MT model. Yet another advantage of such an architecture is that it is in fact a multi-modal and multi-task one: besides speech, it also accepts text input for translation and it produces ASR output either in stand-alone mode or as a side-product of the AST task. 

This multi-input/modal view on the AST task is firmly anchored in the reality of practical applications, so we take it as a fundamental design choice: we aim to build a model that delivers both state of the art ASR and MT performance, while optimizing the AST performance within these constraints. 

Translating ASR 1-best output has the obvious disadvantage that any further training (fine-tuning) on AST parallel data specific to a given domain is unable to back-propagate cross-entropy loss gradient through the interface between the ASR and the MT model. For tighter coupling between ASR and MT modules we follow the approach of \citep{dalmia-etal-2021-searchable} that leverages the 1-best ASR alignment and sends the ASR encoder embeddings aligned with the 1-best ASR sequence to the MT model. This results in a cascade architecture that allows back-propagation gradient to flow from the MT model into the ASR components. 

The ASR model in our work uses a conformer encoder architecture~\citep{gulati2020conformer}, pre-trained on a large amount of speech data as described in the Unified Speech Model (\verb+USM+) work~\citep{zhang2023google}. Due to its remarkable trade-off between simplicity and effectiveness, we compute 1-best labels and the alignment using the Connectionist Temporal Classification (CTC),~\citep{ctc}, guided reduction approach~\citep{wang2022accelerating}. Our experiments show only marginal degradation in word error rate over \verb+RNN-T+ models when training/fine-tuning both on MuST-C V2 AST data~\citep{di-gangi-etal-2019-must}.

The MT model uses a standard encoder-decoder transformer architecture using cross attention between decoder and encoder and self-attention within either (the latter being of course causally masked in the decoder). One slight deviation from a standard implementation is the use of rotary position embeddings~\citep{su2022roformer}. We carry out experiments in two setups: a MT model trained on WMT data only, as well as the MuST-C MT model obtained by incrementally training the WMT model on the MuST-C V2 data.

Our novel contribution is the use of an ``exporter'' layer that is trained under L2-loss to ensure a strong match between ASR embeddings and the MT token embeddings for the 1-best sequence. The ``exporter'' output embeddings are fed directly to the MT model in lieu of 1-best token embeddings, thus guaranteeing that the resulting model performs no worse than the 1-best cascade baseline. Further fine-tuning of the ``exporter'' module alone while keeping the parameters for the ASR and text MT components fixed satisfies the design constraint that we outlined above. We find that we can significantly improve AST performance when using the WMT MT back-end. Alas, this is no longer the case when using the MuST-C MT back-end, showing that the ``exporter''-enhanced AST model performs task adaptation, instead of mitigating the impact of feeding errorful ASR transcriptions to the MT model. 

The approach is useful in scenarios where incremental training of the MT model is not an option and yet we seek to improve AST quality by leveraging (speech, transcription, translated transcription) data. More generally, it offers a promising approach for coupling ASR encoders with immutable text models such at large language models (LLM).

Section~\ref{sec:modeling} details our modeling choices for the ASR and MT components and describes the L2 matcher used to train the ``exporter'' module. Section~\ref{sec:exps} presents our experiments; we conclude with sections outlining related and future work.

\section{Modeling} \label{sec:modeling}

We build ASR models by boot-strapping the speech encoder from \verb+USM+ models trained unsupervised using the BEST-RQ algorithm described in \citep{zhang2023google}. The frame synchronous embeddings produced by the \verb+USM+ encoder are fed to various types of decoders:
\begin{itemize}
    \item \verb+Reduced CTC+, as described in~\citep{wang2022accelerating}: at each frame we pick the top scoring token according to the CTC layer softmax, then apply the CTC ``reduction'' by removing blank symbols and collapsing runs of the same token into one single token instance, aligned with the last frame in the run. The CTC softmax layer and \verb+USM+ encoder are trained using CTC loss: maximize the probability of the correct token sequence by summing over all the CTC alignments for it;
    \item \verb+RNN-T+, as described in~\citep{rao2018exploring}: at each step the decoder consumes an input frame and predicts an alignment edge by joining the \verb+USM+ encoding for the frame and the encoding of the token prefix so far according to a RNN language model (RNN-LM). The LM state is not advanced if the edge has $\epsilon$ label; if the edge has a regular token as label then the LM state is advanced to encode the new prefix obtained by appending the newly predicted token to the old one. During training, computing the probability of the reference token sequence by summing over all \verb+RNN-T+ alignments is simplified by the use of a LM with finite context (N-gram), which enables a dynamic programming solution.
    \item \verb+LAS+, as described in~\cite{chan2015listen}: a RNN LM (causally masked transformer model in our case) augmented with cross-attention~\citep{bahdanau2016neural} over the \verb+USM+ encoder output is used to decode the token sequence. The model is trained under cross-entropy loss against the token sequence derived from the correct transcription. Among the downsides of \verb+LAS+ modeling we can mention that there is no natural frame level alignment for the output token sequence and it does not easily support a streaming decoding algorithm. On the upside, it offers a stronger LM in the decoder and more powerful integration of encoder output in the decoder due to soft cross-attention. The latter aspects make the \verb+LAS+ architecture the natural candidate for direct AST modeling.
\end{itemize}

The \verb+USM+ model uses 24 conformer layers of dimension 1024, with a convolution kernel of size 5 for a total of 600M parameters. The \verb+RNN-T+ LM is an LSTM with two layers, input/output embedding of dimension 128 and 640 hidden dimension. The \verb+LAS+ decoder transformer has 6 layers of dimension 512 and 2048 hidden dimension, 8 attention heads and rotary position embedding. For training we use the Adam optimizer with transformer learning rate schedule and exponential moving average. Decoding is done using beam search of size 8.

The MT models are standard transformer architectures consisting of 18 encoder and 6 decoder layers of dimension 1024, using 16 attention heads and rotary position embedding, resulting in about 300M parameters. Optimization uses sharded adafactor with Adam decay schedule (0.999 decay, 0.9 beta1, 5.0 gradient norm clipping), linear ramp-up and exponential decay learning rate schedule. Training uses exponential moving average with weight 0.999, label smoothing and dropout for regularization.

Translating ASR 1-best output has the obvious disadvantage that any further training (fine-tuning) on AST parallel data specific to a given domain is unable to back-propagate cross-entropy loss gradient through the interface between the ASR and the MT model. For tighter coupling between ASR and MT modules we follow the approach of \citep{dalmia-etal-2021-searchable} that leverages the 1-best ASR alignment and sends the ASR encoder embeddings aligned with the 1-best ASR sequence to the MT model. Other approaches to accomplish this exist, see Section~\ref{sec:related_work}.

\subsection{L2 Matcher for Cascade AST}

Assuming that the SPM (Sentence Piece Model),~\cite{kudo-richardson-2018-sentencepiece}, used for ASR is the same as the one used for source sentence in the MT model, we embed the \verb+USM+ encodings aligned with the ASR output tokens such that they match the MT input embeddings. This is accomplished by an ``exporter'' module (3 conformer layers) that re-embeds the \verb+USM+ encodings. The exporter output is optionally passed through a linear layer that ensures dimensionality match with the MT input token embeddings. All layers in the ``exporter'' module are trained using L2 loss on ASR parallel data consisting of (speech, transcription) pairs while keeping all the other parameters (MT token embeddings, ASR \verb+USM+ encoder and CTC layer) fixed. The purpose of this first training stage is to approximate the hard cascade baseline. A second (optional) stage of training further estimates the ``exporter'' parameters on AST data, while keeping the \verb+USM+ and MT modules frozen.
\begin{figure*}
    \centering
    \fbox{
    \includegraphics[width=12cm]{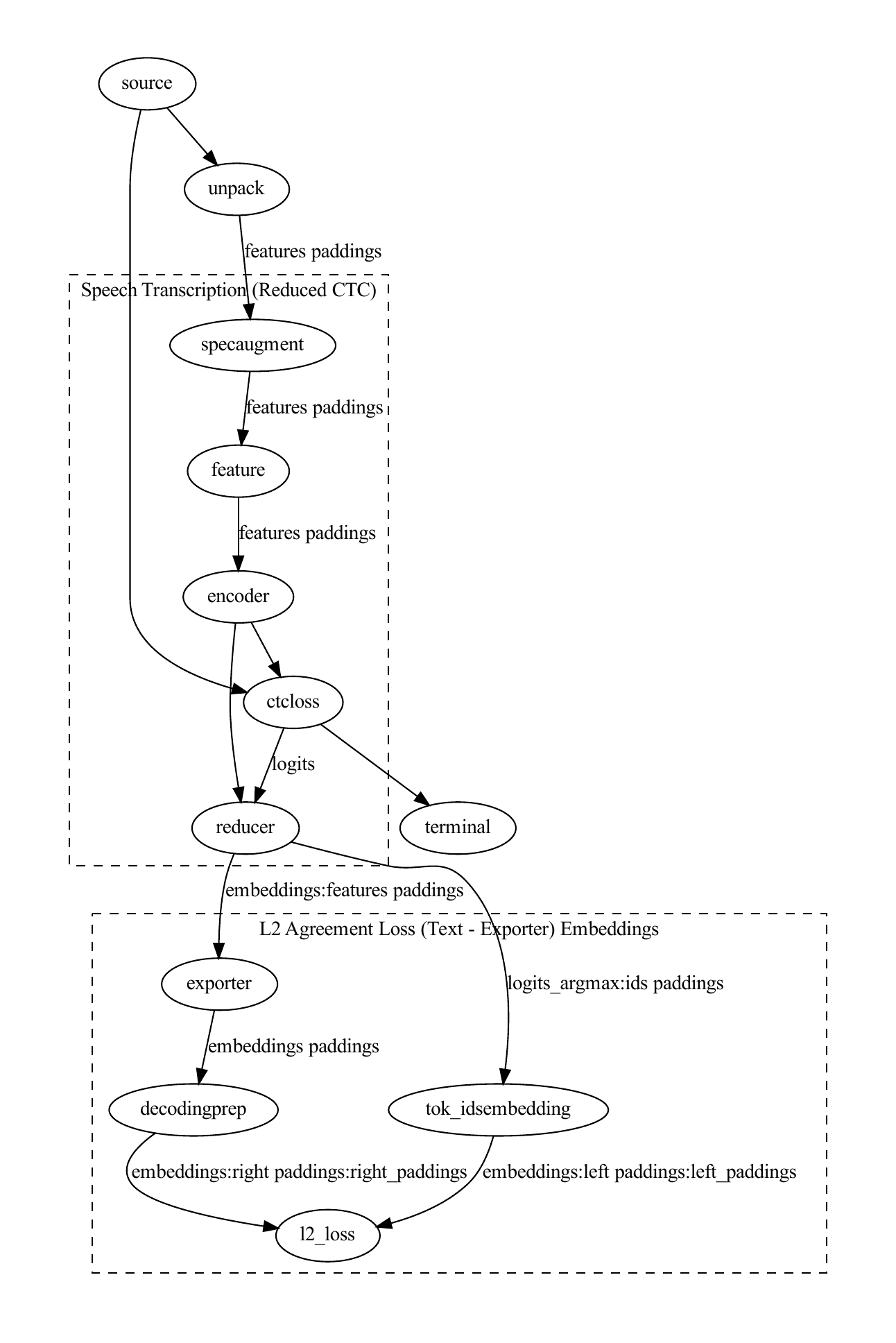}  
    }
    \caption{Graph model description for training the L2 loss matcher.}
    \label{fig:l2}
\end{figure*}

Once this is accomplished we build a cascade AST model that sends the ``exporter'' output straight to the transformer encoder of the MT model, bypassing the MT input token embeddings.  This results in a cascade architecture that allows back-propagation gradient to flow from the MT model into the ASR components. We present experiments showing that:
\begin{itemize}
    \item L2 loss initialization of the exporter does indeed match 1-best cascade AST results
    \item further training the exporter layer parameters on parallel AST data consisting of (speech, translation) pairs can improve AST performance while keeping ASR and text MT capability intact since those model parameters are not updated.
\end{itemize}
\begin{figure*}
    \centering
    \fbox{
    \includegraphics[width=12cm]{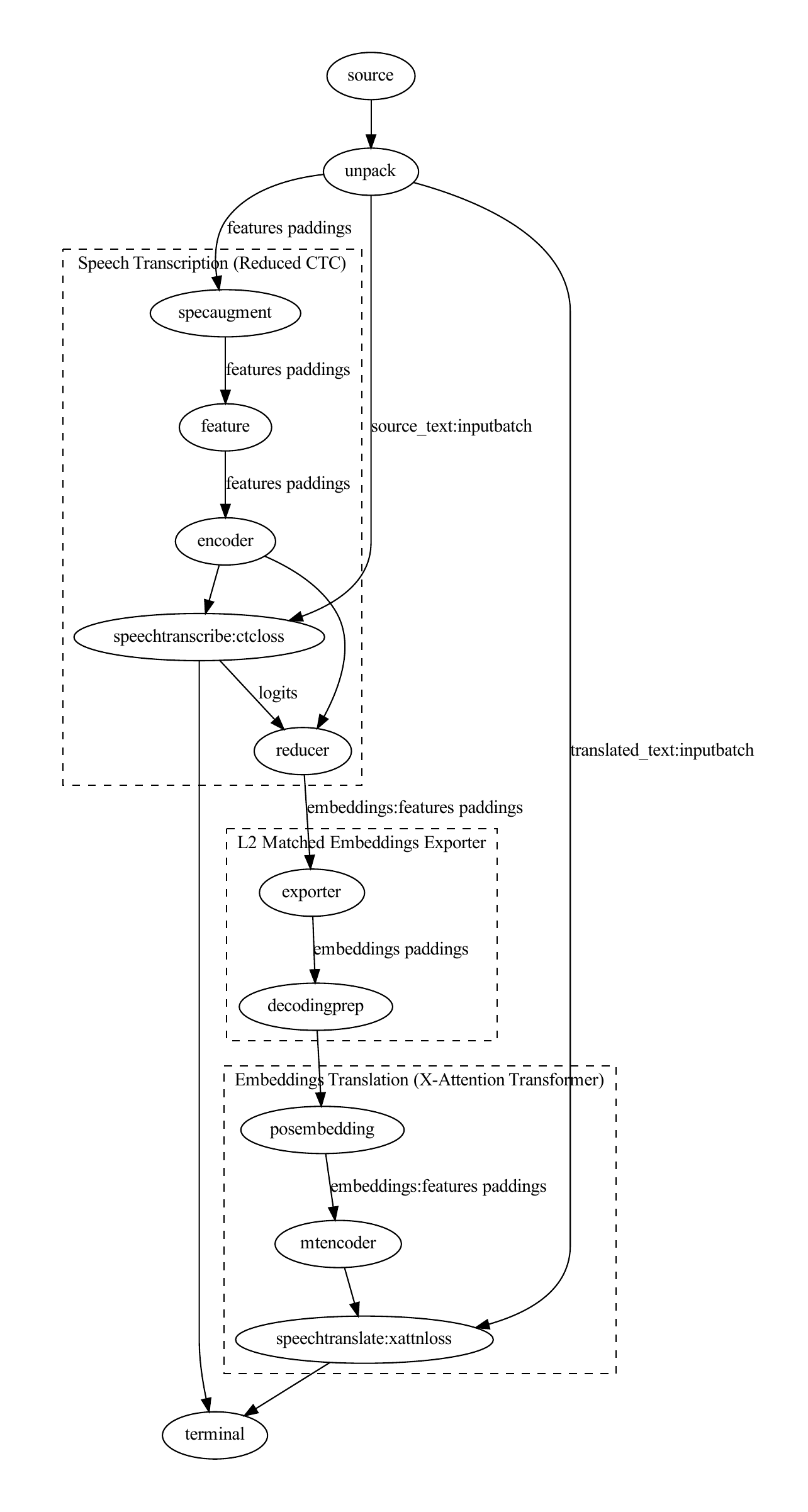}  
    }
    \caption{Graph model description for using the L2 loss matcher in cascade AST.}
    \label{fig:ls_ast}
\end{figure*}

\section{Experiments} \label{sec:exps}

We carried out AST experiments on the En-De subset of the MuST-C~\citep{di-gangi-etal-2019-must} V2 data. We did not pre-process the transcriptions or translations in any way, thus working in the ``written text'' domain. This choice was mostly due to the fact that MT models are trained on such text data; it also makes our work easily reproducible. The ASR model thus needs to generate all the extra information that may not be available in the speech signal such as punctuation, capitalization, correct formatting of numbers, dates, etc. As a result, the word error rate (WER) results listed in Table~\ref{table:wer} may seem higher than expected.

The text data is tokenized using sentence piece models (SPM,~\cite{kudo-richardson-2018-sentencepiece}) built from the MuST-C data separately for the source and target language. Unless specified otherwise we use byte-pair encoding SPM with a vocabulary size of 1024, and a short list of user defined symbols specific to the MuST-C data such as: \verb+(Laughter), (Applause), (Sobs), (Music), (Mock sob), (Piano music)+, \verb+(Sighs), (Singing), (Sings), (Whistling)+ and their German counterparts. The output vocabulary of the MT model used for AST was 32768; the MT baseline models use a SPM vocabulary of 32678, however that was reduced to 1024 for the AST cascade experiments to match the ASR vocabulary size. We did not observe a significant impact in SacreBleu due to a smaller source SPM vocabulary.

Speech encoders are initialized using the \verb+USM+ model trained unsupervised on about 12M hours of speech data using the BEST-RQ algorithm described in \citep{zhang2023google}. The \verb+WMT+ translation model is trained on data available for the \verb+WMT'21+ constrained evaluation setup containing approximately 90M EnDe sentence pairs. That is to be compared with the 400hrs of EnDe speech and 251k sentence pairs available in the MuST-C V2 training set. 

\subsection{ASR experiments}

The ASR models used in our experiments are trained incrementally from \verb+USM+~\citep{zhang2023google} models. We started with a 600M parameter model and finetuned both \verb+RNN-T+ and \verb+Reduced CTC+ ASR models on the MuST-C data. The results are presented in Table~\ref{table:wer}.
\begin{table}[ht]
    \centering
    \begin{tabular}{|l|r|r|r|}\hline
    Model         &  \multicolumn{3}{c|}{WER (\%)}\\
                  & dev  & tst-COMMON  & tst-HE \\\hline
    \verb+RNN-T+         & 9.2  &  9.6        & 7.4    \\\hline
    \verb+Reduced CTC+   & 10.2 & 10.2        & 8.0   \\\hline
    \end{tabular}
    \caption{Word Error Rate of RNN-T and reduced CTC models built by fine-tuning USM models on En-De MuST-C data.}
    \label{table:wer}
\end{table}

\subsection{MT experiments}

A first set of MT experiments evaluated the performance of the \verb+WMT+ model on \verb+MuST-C+ before and after incremental training. In addition to using the correct ASR transcript we also experimented with training and/or evaluating on the 1-best transcript produced by the \verb+RNN-T+ ASR model. The results are presented in Table~\ref{table:mt}.  Replacing correct transcription for the source sentence with the \verb+RNN-T+ 1-best degrades performance by 2-3 SacreBleu, depending on the evaluation data set.


\begin{table}[ht]
    \centering
    \begin{tabular}{|l|r|r|r|r|}\hline
    Model                 & Source           &  \multicolumn{3}{c|}{SacreBleu (\%)}\\
                          & Transcript       & dev   & tst-COMMON  & tst-HE  \\\hline
    \verb+WMT+         & correct          & 32.7  & 33.8        & 31.6    \\
    \verb+WMT-MuST-C+  & correct          & 35.9  & 36.4        & 36.2    \\\hline
    \verb+WMT+         & \verb+RNN-T+ 1-best     & 31.1  & 30.8        & 29.9    \\
    \verb+WMT-MuST-C+  & \verb+RNN-T+ 1-best     & 33.8  & 33.0        & 33.7    \\\hline
    \end{tabular}
    \caption{SacreBleu of WMT baseline and incrementally trained WMT-MuST-C MT model on MuST-C data.}
    \label{table:mt}
\end{table}

We have also incrementally trained the \verb+WMT+ model on MuST-C data where we replaced the correct source transcription with the ASR 1-best but we did not observe any change relative to the model trained using the correct transcription. This is a surprising result, since we were expecting a slight improvement when evaluating on matched data containing \verb+RNN-T+ 1-best transcriptions. It seems that during incremental training the translation model is able to integrate over the noise introduced by ASR errors and the only reason for degradation in a cascade AST architecture is errorful transcription of source speech at inference/test time.

\subsection{AST Experiments}

The very first AST experiment was to compare direct model performance with that of the cascade model. In earlier experiments with written text ASR carried out on LibriTTS~\cite{zen2019libritts} data we observed slightly better performance of \verb+LAS+~\citep{chan2015listen} versus \verb+RNN-T+ models~\citep{rao2018exploring} built by fine-tuning a \verb+USM+ encoder: 8.5\% and 10.9\% on \verb+dev-clean+ and \verb+test-other+ partitions versus 8.8\% and 11.2\%. Given that observation and the fact that the MT decoder relies heavily on cross-attention to reorder the target side tokens relative to the source, we build a direct AST model using the \verb+LAS+ architecture. Table~\ref{table:direct_ast} compares SacreBleu for the direct AST model with the cascade model built by feeding \verb+RNN-T+ 1-best transcriptions to the MT model. The \verb+RNN-T+ model is built by fine-tuning a \verb+USM+ encoder (600M parameters) on MuST-C data. The same approach is used to build the \verb+LAS+ direct AST model.
\begin{table}[ht]
    \centering
    \begin{tabular}{|l|r|r|r|r|}\hline
    Model                 & Source Transcript&  \multicolumn{3}{c|}{SacreBleu (\%)}\\
                          &                  & dev   & tst-COMMON  & tst-HE  \\\hline
    \multicolumn{2}{|l|}{Direct model: USM encoder + LAS decoder} & 28.2 & 27.2 & 27.3 \\\hline
    \verb+WMT+         & \verb+RNN-T+ 1-best     & 31.1  & 30.8        & 29.9    \\
    \verb+WMT-MuST-C+  & \verb+RNN-T+ 1-best     & 33.8  & 33.0        & 33.7    \\\hline
    \end{tabular}
    \caption{Comparison of direct AST model with cascade architecture using either WMT baseline or incrementally trained WMT-MuST-C MT model on MuST-C data together with RNN-T ASR model also incrementally trained on MuST-C data.}
    \label{table:direct_ast}
\end{table}

It is safe the attribute the large difference of 5-6\% SacreBleu between direct and cascade models to the disparity in training data highlighted at the beginning of section~\ref{sec:exps}: due to their modular approach, cascade models have access to a much larger amount of training data in addition to what is available in \verb+MuST-C+. This situation is quite representative of practical scenarios, so we focused our next set of experiments on cascade models. As explained in section~\ref{sec:intro}, we take a multi-task and multi-input view on the AST problem: ASR transcripts are a side product of the speech translation task, and we seek to optimize speech translation performance while preserving state-of-the-art performance for both ASR and text MT.

\subsubsection{L2 Matcher Experiments}

A first experiment trained the exporter component (3 layer conformer) to ensure L2 match with the MT token embeddings. This uses the \verb+MuST-C+ parallel data (speech, transcription). Table~\ref{table:l2_loss} shows the loss values after 25,000 training steps using a global batch size of 128. A very good fit between the encodings exported after CTC reduction and the MT token embeddings is achieved, within 0.001 (10/1024) absolute diff per dimension.
\begin{table}[ht]
    \centering
    \begin{tabular}{|l|r|r|r|}\hline
    MT Model              &  \multicolumn{3}{c|}{L2-loss/token}\\
                          & dev   & tst-COMMON  & tst-HE  \\\hline
    \verb+WMT+            & 9.0   & 9.2         & 8.9     \\
    \verb+WMT-MuST-C+     & 9.5   & 9.8         & 9.4     \\\hline
    \end{tabular}
    \caption{L2 loss per token summed across 1024 dimensions.}
    \label{table:l2_loss}
\end{table}

We then proceeded to train the exporter from this initial point on parallel (speech, translation) MuST-C data in an attempt to improve AST performance beyond the initial one (step 0). We also verified that the AST performance at step 0 matches the cascade 1-best one as shown in Table~\ref{table:l2_loss_ast_wmt}.


  
\begin{table}[ht]
    \centering
    \begin{tabular}{|l|r|r|r|}\hline
    Cascade AST Model     &  \multicolumn{3}{c|}{SacreBleu}\\
                          & dev   & tst-COMMON  & tst-HE  \\\hline
    transcript            & 32.7  & 34.2        & 30.9    \\
    \verb+RNN-T+ 1-best          & 31.1  & 30.8        & 29.9    \\\hline
    \verb+Reduced CTC+ 1-best    & 30.6  & 30.8        & 28.5    \\
    \verb+Reduced CTC+ L2 loss, step 0    & 30.7  & 30.6        & 28.7    \\
    \verb+Reduced CTC+ L2 loss, step 8k   & 32.5  & 32.5        & 31.8    \\\hline
    \end{tabular}
    \caption{AST performance for cascade architecture using L2 loss matcher for reduced CTC ASR, WMT MT model.}
    \label{table:l2_loss_ast_wmt}
\end{table}
As mentioned already, we first observe that the \verb+Reduced CTC+ 1-best AST performance is very close to the \verb+RNN-T+ 1-best one, owing to the small difference in WER between the two ASR models. The \verb+Reduced CTC+ L2 loss architecture matches it at step 0, as expected due to the very close match in L2 loss between exporter encodings and \verb+WMT+ token embeddings. Finally, we were pleasantly surprised to see that further training of the \verb+Reduced CTC+ L2 loss model improves performance significantly, approaching the \verb+WMT+ model performance on correct transcripts. 

We wish to note that when training the exporter on MuST-C AST data, there is task specific ``leakage'' into the model despite the fact that ASR/MT parameters are frozen. That should explain the ability to beat the \verb+WMT+ translation model performance when feeding it correct transcripts for the MuST-C data, as observed on the tst-HE test set (last row in Table~\ref{table:l2_loss_ast_wmt}).

A second set of experiments confirmed that this was indeed the main reason for the improvements above: when using the \verb+WMT-MuST-C+ translation back-end (a \verb+WMT+ model adapted on MuST-C data) and an exporter matched to the token embeddings of this model there is no longer a significant gain from further training of the \verb+Reduced CTC+ L2 loss model, as shown in Table~\ref{table:l2_loss_ast_incr}.
\begin{table}[ht]
    \centering
    \begin{tabular}{|l|r|r|r|}\hline
    Cascade AST Model     &  \multicolumn{3}{c|}{SacreBleu}\\
                          & dev   & tst-COMMON  & tst-HE  \\\hline
    transcript            & 35.3  & 35.8        & 35.3    \\
    \verb+RNN-T+ 1-best          & 33.8  & 33.0        & 33.7    \\\hline
    \verb+Reduced CTC+ 1-best    & 32.8  & 32.0        & 32.4    \\
    \verb+Reduced CTC+ L2 loss, step 0    & 32.4  & 32.3        & 33.1    \\
    \verb+Reduced CTC+ L2 loss, step 11k  & 32.7  & 32.3        & 32.5    \\\hline
    \end{tabular}
    \caption{AST performance for cascade architecture using L2 loss matcher for Reduced CTC ASR, incrementally trained WMT-MuST-C MT model.}
    \label{table:l2_loss_ast_incr}
\end{table}

A last set of experiments aim at understanding to what extent the L2 loss initialization of the exporter module is useful. Table~\ref{table:l2_loss_ablation} shows the results of ablation experiments using both \verb+WMT+ and \verb+WMT-MuST-C+ translation models: we use random initialization for the exporter module instead of the L2 loss one described above.
\begin{table}[ht]
    \centering
    \begin{tabular}{|l|l|r|r|r|}\hline
    MT Model & Exporter Initialization & \multicolumn{3}{c|}{SacreBleu}\\
               &                       & dev   & tst-COMMON  & tst-HE  \\\hline
    \verb+WMT+        & random                & 31.2  & 30.3        & 30.8    \\
    \verb+WMT+        & L2 loss matcher       & 32.5  & 32.5        & 31.8    \\\hline
    \verb+WMT-MuST-C+ & random                & 31.0  & 30.4        & 31.3    \\
    \verb+WMT-MuST-C+ & L2 loss matcher       & 32.7  & 32.3        & 32.5    \\\hline
    \end{tabular}
    \caption{Ablation experiments comparing AST performance for cascade architecture using exporter initialized randomly versus L2 loss matcher with both WMT and incrementally trained WMT-MuST-C MT model.}
    \label{table:l2_loss_ablation}
\end{table}
Using the L2 loss matcher to initialize the exporter before further training on the MuST-C data set is beneficial in both experimental conditions.

\section{Related Work} \label{sec:related_work}

A similar approach of coupling the ASR and MT components in cascade AST models, as well as their performance relative to direct models when leveraging all the data available is presented in~\citep{bahar2020tight}. Coupling the ASR and MT model using the ASR model embeddings along the 1-best alignment was introduced in~\cite{dalmia-etal-2021-searchable}; we take the additional step of adding the exporter module and initialize it to match the MT model token embeddings under the L2 loss. As shown by the ablation experiments, this initialization is important for achieving the best AST performance. Our work follows the same ideas as in LegoNN~\citep{dalmia2022legonn}, primarily by building modular models that can leverage significantly larger amounts of training data than the limited parallel data available to build AST models; it stems from the same project as~\citep{lego_features}. The use of CTC reduced ASR for AST is novel to the best of our knowledge.

\section{Conclusions and Future Work}

We have presented a modular approach to building cascade AST models that guarantees that the resulting model performs no worse than the 1-best cascade baseline while preserving state-of-the-art ASR and text MT performance for a given task. The approach could be useful in scenarios where incremental training of the MT model is not an option and yet we seek to improve AST quality by leveraging (speech, transcription, translated transcription) data.

Future work could improve our approach in a few directions. One is to replace the CTC reduced ASR with \verb+RNN-T+: since the latter also produces a frame-level alignment one could export the \verb+RNN-T+ encoder embeddings just as well, guarantee even better ASR 1-best performance which is directly correlated with AST quality. Another is to augment the MT encoder with a (perhaps gated) cross-attention mechanism and allow it to reach into the encoder embeddings sequence as needed to extract any complementary information that may improve AST performance. The challenge is of course maintaining modularity and ensuring that text translation performance is not impacted negatively. It would be interesting to compare the two stage training approach with a single joint optimization, where the L2 loss is used as a regularizer rather than a pretraining stage

Perhaps even more promising is a scenario that uses the matched-embeddings approach to couple ASR with an immutable LLM, leveraging long contextual information in refining the typically short segment-level (utterance) ASR results.

\section{Acknowledgements}
Thanks to the Pax~\citep{pax_github} developers and users in Google for help in setting up experiments and debugging, in particular Laurent El Shafey; Markus Freitag for providing the WMT training data; Maxim Krikun and Ankush Garg for providing a Pax MT model implementation that we could build on; Sid Dalmia for useful comments and pointers to literature; Aren Jansen for a careful review of the paper and suggestions for potential improvements as future work. 

\bibliography{ast}
\bibliographystyle{acl_natbib}
\end{document}